\documentclass[letterpaper, 10 pt, conference]{ieeeconf}  

\IEEEoverridecommandlockouts                              

\usepackage{graphicx} 
\usepackage{epsfig} 
\usepackage{amsmath} 
\usepackage{float}
\usepackage{caption}
\usepackage{array, multirow, multicol}
\usepackage{algpseudocode}  
\usepackage{algorithm}
\usepackage{relsize}
\usepackage{amsmath,mathpazo}
\usepackage{color}
\usepackage{ragged2e}
\usepackage{tabularx}
\usepackage{cite}
\usepackage{hyperref}

\definecolor{matt_color}{rgb}{.7,0,.7}
\definecolor{charlie_color}{rgb}{1,0,0}
\definecolor{patrick_color}{rgb}{0,0.7,0}
\definecolor{youliang_color}{rgb}{0,0.3,0.7}

\newcommand{\comment}[1]{} 

\newcommand{\matt}[1]{\textcolor{matt_color}{\comment{matt: #1}}}

\newcommand{\patrick}[1]{\textcolor{patrick_color}{\comment{patrick: #1}}}
\newcommand{\youliang}[1]{\textcolor{youliang_color}{\comment{youliang: #1}}}

\title{\LARGE \bf
Stretch with Stretch: Physical Therapy Exercise Games Led by a Mobile Manipulator
}

\author{Matthew Lamsey$^{1}$, You Liang Tan$^{1}$, Meredith D. Wells$^{2}$, Madeline Beatty$^{1}$, Zexuan Liu$^{1}$, \\ Arjun Majumdar$^{1}$, Kendra Washington$^{1}$, Jerry Feldman$^{3}$, Naveen Kuppuswamy$^{4}$, \\ Elizabeth Nguyen$^{2, 5}$, Arielle Wallenstein$^{2}$, Madeleine E. Hackney$^{2}$, and Charles C. Kemp$^{1}$%
\footnotemark{}
}

\newenvironment{first_caption}
  {\par}
  {\par\addvspace{\bigskipamount}}

\begin{document}

\twocolumn[{%
\renewcommand\twocolumn[1][]{#1}%
\maketitle
\thispagestyle{empty}
\pagestyle{empty}

\begin{center}
    \centering
    \vspace{-6mm}
    \includegraphics[width=1.0\linewidth]{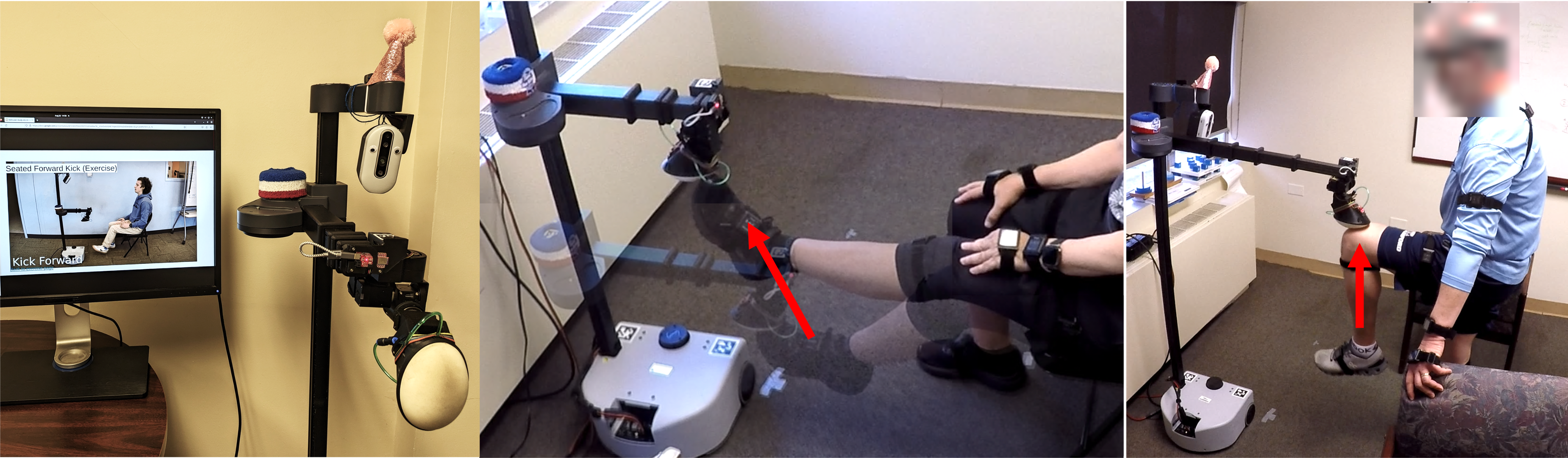}
    \label{fig:banner}
\end{center}
\vspace{-3mm}

\begin{first_caption}
 Fig. 1.  (\textbf{Left}) Stretch with Stretch (SWS), a system for personalized robot-led physical therapy. Video instructions are shown on an adjacent screen. (\textbf{Middle}) SWS sampling the range of motion of a participant with Parkinson's disease (PD). (\textbf{Right}) A participant with PD performing a standing high knees exercise.
\end{first_caption}
\vspace{-2mm}
}]
\setcounter{figure}{1}  
\setcounter{footnote}{1}

\footnotetext{The Institute for Robotics and Intelligent Machines at the Georgia Institute of Technology (GT). $^{2}$Emory University School of Medicine. $^{3}$The Parkinson's Foundation. $^{4}$Toyota Research Institute (TRI). $^{5}$University of Texas Long School of Medicine. This work is supported by TRI grant GR00010714 and the McCamish Parkinson's Disease Innovation Program Blue Sky grant DE00021631. Charles C. Kemp was an associate professor at GT when contributing to this research. He now works full-time for Hello Robot Inc., which sells the Stretch RE1.}%


\begin{abstract}
Physical therapy (PT) is a key component of many rehabilitation regimens, such as treatments for Parkinson's disease (PD). However, there are shortages of physical therapists and adherence to self-guided PT is low. Robots have the potential to support physical therapists and increase adherence to self-guided PT, but prior robotic systems have been large and immobile, which can be a barrier to use in homes and clinics. We present Stretch with Stretch (SWS), a novel robotic system for leading stretching exercise games for older adults with PD. SWS consists of a compact and lightweight mobile manipulator (Hello Robot Stretch RE1) that visually and verbally guides users through PT exercises. The robot's soft end effector serves as a target that users repetitively reach towards and press with a hand, foot, or knee. For each exercise, target locations are customized for the individual via a visually estimated kinematic model, a haptically estimated range of motion, and the person's exercise performance. The system includes sound effects and verbal feedback from the robot to keep users engaged throughout a session and augment physical exercise with cognitive exercise. We conducted a user study for which people with PD ($n=10$) performed 6 exercises with the system. Participants perceived the SWS to be useful and easy to use. They also reported mild to moderate perceived exertion (RPE). 


\end{abstract}

\section{Introduction}

An aging global population \cite{world2015world} is increasing the burden on healthcare systems \cite{coughlin2006old}. Additionally, approximately one million people live with Parkinson’s disease (PD) in the United States \cite{yang2020current}. People with PD have mobility impairments, particularly bradykinesia, hypometria, and initiation of movement deficits \cite{factor2014freezing, pal2016global, stegemoller2014timed}. Many people with PD also have cognitive impairments, including deficits in set switching and challenges with dual tasking \cite{domellof2015cognitive, geurtsen2014parkinson, litvan2011mds, pretzer2009parkinson}. Physical and occupational therapists frequently focus on treating these impairments with tailored and progressive exercises \cite{o2014clinical, salgado2013evidence, schenkman2012exercise}, which improve quality of life and functional ability and may slow disease progression \cite{caglar2005effects, ferrazzoli2016does, van2013effects, mak2017long}. However, adherence to therapeutic exercise regimens remains difficult for many patients to achieve due to factors including a physical therapist workforce shortage \cite{buerhaus2008current} and deficits in intrinsic motivation \cite{pickering2013self}. In addition, the dose of a given exercise which is necessary to evince successful results often exceeds that which a therapist can individually provide even within small group settings \cite{clarke2016physiotherapy}. These factors result in lost opportunities for improved health.

Traditional physical therapy (PT) often involves one-on-one guided exercises with a therapist; however, robotic systems are increasingly being integrated into therapeutic protocols. Drawing inspiration from these systems, we introduce Stretch with Stretch (SWS): a physically and socially interactive robotic system for leading stretching exercise games for older adults with PD. Our platform consists of a Hello Robot Stretch RE1 mobile manipulator \cite{kemp2022design} with a soft-bubble end effector \cite{kuppuswamy2020soft}. SWS uses a novel model for personalized exercise to plan and lead particpants through repetitive stretching exercises that are based on a user's body dimensions, a user's range of motion, and the user's performance throughout a session. These exercises directly address motor impairments (e.g. hypometria) associated with PD. SWS also includes cognitive exercises to manage difficulty with dual-tasking \cite{kalyani2019effects, smith2018communication}, a common PD symptom. Together, these features aim to keep users engaged throughout an exercise session.

We evaluated our system through a user study ($n=10$) of people with Parkinson's disease (PWP). We found that PWP perceived the system to be useful and easy to use, and they perceived the intensity of the exercise session to be mildly to moderately difficult. These findings suggest that our system is a viable option for the treatment of PD in a clinical setting.

In summary, our contributions follow:

\begin{itemize}
\item We present a novel robotic system for physical therapy comprised of an off-the-shelf mobile manipulator equipped with a soft-bubble end effector.
\item We present a novel model for exercise that defines personalized locations for repetitive physical contact by a user's body based on an exercise specific kinematic model, haptic estimation of range of motion, and the user's performance. 
\item We present 10 robot-led exercises for PWP that can be combined with a cognitive exercise. 
\item Through a user study ($n=10$) involving PWP performing 6 exercises, we show that PWP perceive SWS as useful and easy to use. PWP reported the exercise as requiring mild to moderate perceived exertion.
\end{itemize}

\section{Related Work}

Robots have been used to lead many types of physical therapy exercises through physical interactions with users, such as dancing \cite{chen2015evaluation, chen2017older}, hand-clapping exercise games \cite{fitter2020exercising}, post-stroke rehabilitation \cite{johnson2003design, krebs2004rehabilitation}, walking rehabilitation \cite{wuversatile, regmi2020design}, and motor-cognitive rehabilitation \cite{aprile2020robotic}. However, these systems often do not offer a significant level of personalization of the difficulty of the therapeutic exercises, and they typically include purpose-built hardware that is only suitable for use in a clinic due to size, cost, and complexity. Socially assistive robotic (SAR) systems have also been deployed for limited applications including seated stretching exercises \cite{fasola2012using} and tabletop rehabilitation games \cite{feingold2021robot}, but these systems are limited by their lack of physical interaction with users. In the context of PD, exercise games have proven to be a successful treatment method \cite{barry2014role}, indicating the possibility of creating an interactive, game-based robotic therapy system specifically for PD management.

The use of Virtual Reality (VR) and Augmented Reality (AR) systems for PT has also been investigated. VR systems, which include screen-based systems such as the Nintendo Wii \cite{holmes2013effects} as well as head-mounted display systems \cite{campo2022wearable, sanchez2020impact}, have been used for PD rehabilitation and result in short-term improvements in motor performance \cite{kwon2023systematic}. However, the majority of this work has focused on gait and balance training, rather than range of motion training. Other limitations of VR systems include inaccuracies in human kinematic tracking, which lead to erroneous feedback \cite{demers2020kinematic} and possible motion sickness from use \cite{saredakis2020factors}. Previous studies of AR treatments for PD involved head-mounted AR devices that showed motivational exercise videos \cite{tunur2020augmented} and movement cues to prevent freezing of gait \cite{janssen2020effects}. While the AR system described in \cite{tunur2020augmented} was safe and rated as enjoyable, neither system caused significant positive changes to participants' health. Comparisons between VR and robots yielded mixed results; \cite{cohavi2022young} found that users preferred a VR system over an SAR for gameified cognitive training tasks, while \cite{tapus2009role} found that users preferred interacting with a real robot versus a simulated robot.

\section{Robotic System}

\subsection{Robotic Platform for Exercise}
\begin{figure}
    \centering
    \includegraphics[width=0.475\textwidth]{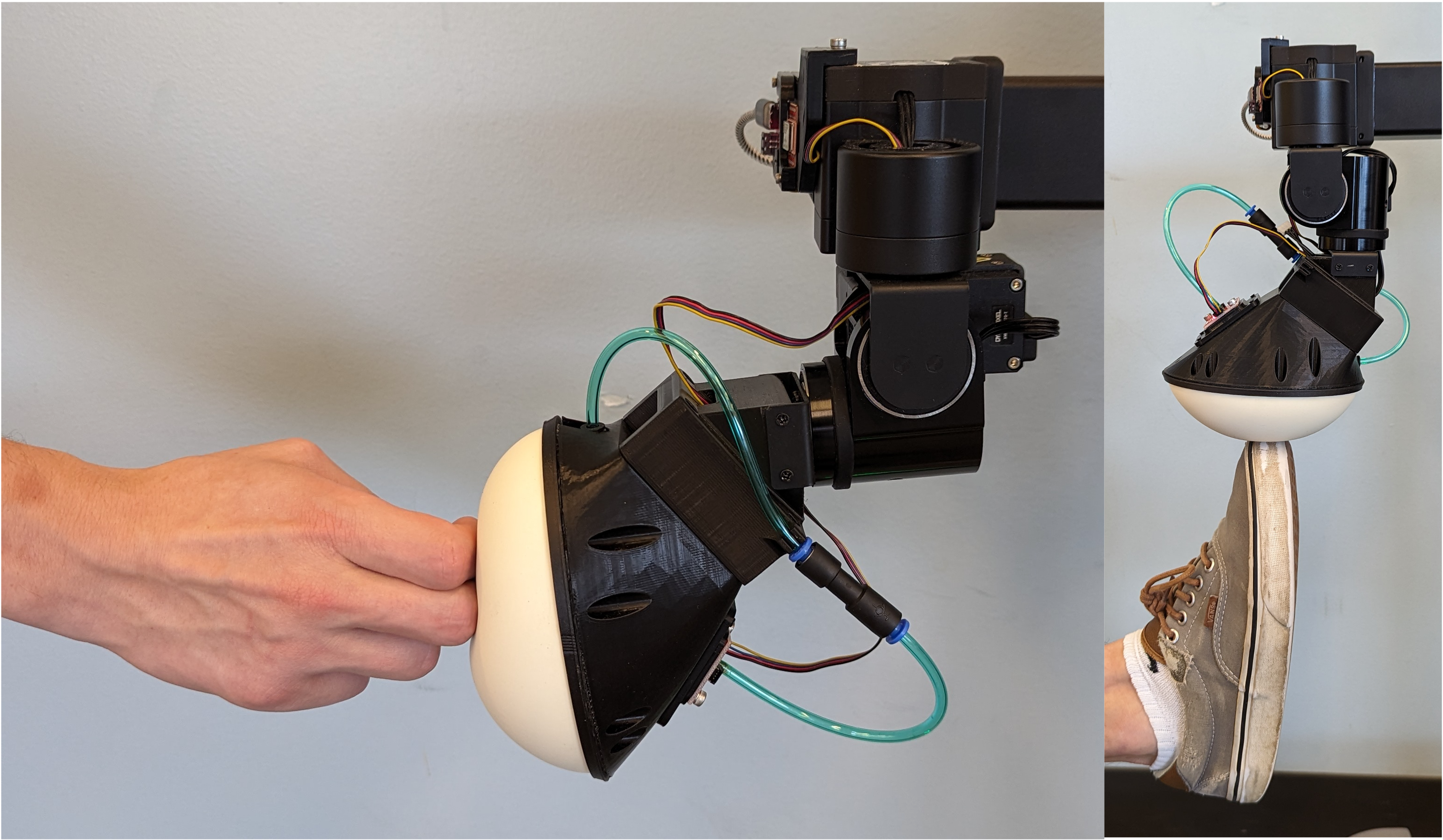}
    \caption{Exercises are led by a Hello Robot Stretch RE1 \cite{kemp2022design} with a soft bubble end effector \cite{kuppuswamy2020soft}. Users press the bubble with different body parts.
    } \vspace{-3mm}
    \label{fig:zeste}
\end{figure}

\patrick{about this section overall: more details about the robot, your modifications to it, less about the audio. Dividing into more clear sections may help you do this}
We use a modified Hello Robot Stretch RE1 \cite{kemp2022design} to lead stretching exercises. The robot's stock end effector is replaced by a pressurized soft-bubble paw \cite{kuppuswamy2020soft}, which serves as the target for users to touch with different parts of their body (Figure \ref{fig:zeste}). Contact between the bubble and body parts can be identified by measuring the change in internal bubble pressure. \patrick{The bubble gripper deserves its own subsection. How does it work? can you make a diagram of the cross-section? what is the minimum force it can detect? It allows the robot to detect punches, kicks, taps, etc} \matt{the sensitivity of the bubble is set to +250Pa from bubble pressure at startup. This doesn't always map to the same force because it depends on where you hit the bubble. It also depends on the individual bubble; they all qualitatively feel like they have different firmnesses b/c the latex wears out over time.}



The Stretch RE1 robot consists of a mobile base and an extensible arm to reach over a large volume while remaining portable and suitable for home use. This is distinct from prior robotic systems, which have been large, heavy, and stationary \cite{feingold2021robot, fitter2020exercising}. To highlight the limitations of stationary robots for this application, we simulated a Rethink Robotics Baxter (as used in \cite{fitter2020exercising}) reaching the actual target locations achieved by SWS during its deployment. We define a reachable point as a location that the robot's end effector can reach within 0.02m with any orientation. We performed two simulations: one with Baxter's base placed at the same location as SWS's starting location during SWS, and one with Baxter's base position numerically optimized using CMA-ES \cite{hansen2003reducing} to maximize the number of reachable target locations.

\begin{figure}
    \centering
    \includegraphics[width=0.475\textwidth]{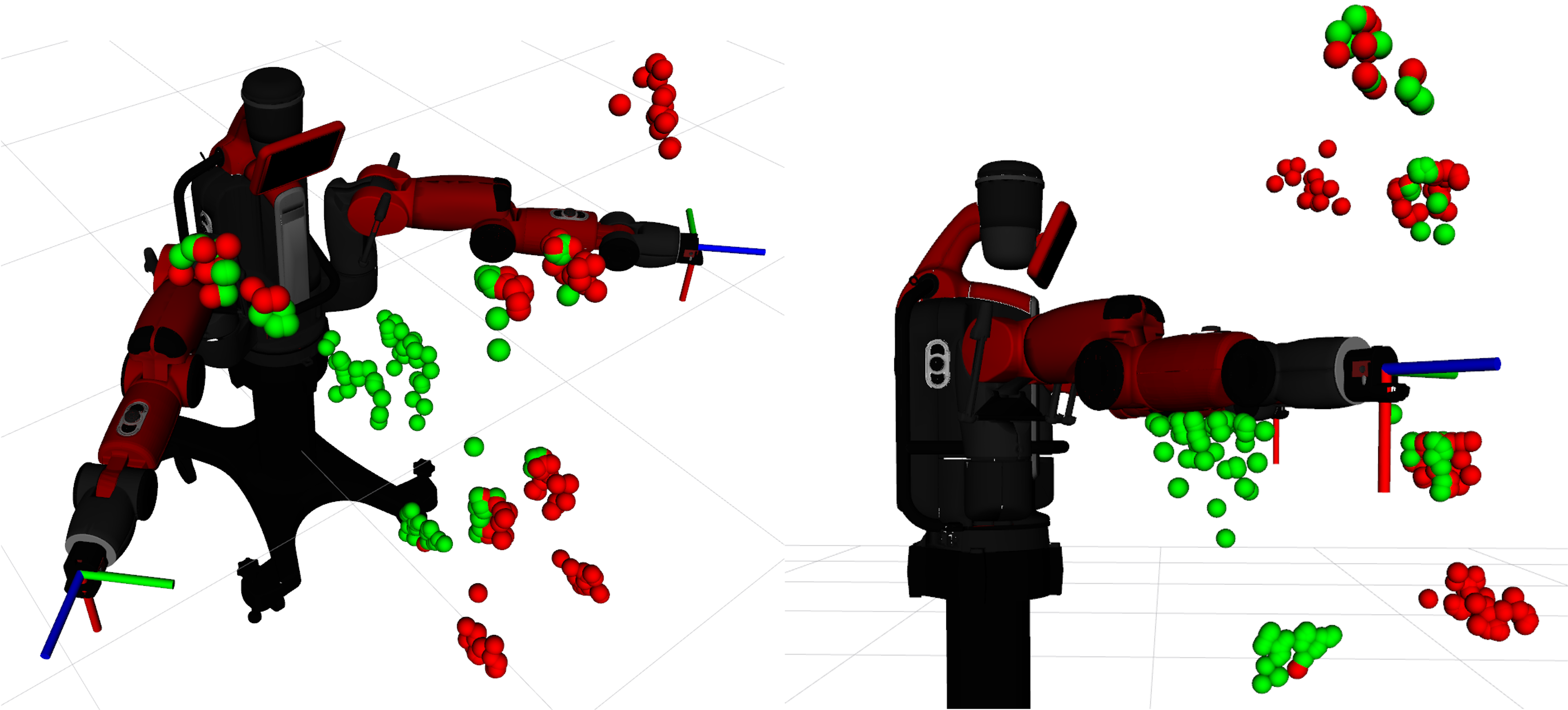}
    \caption{Actual SWS target locations colored by whether they can be reached (green) or not (red) by a Rethink Robotics Baxter (as used in \cite{fitter2020exercising}).} \vspace{-3mm}
    \label{fig:reachability}
\end{figure}

The results from the first simulation are shown in Figure \ref{fig:reachability}. Baxter is only able to reach about 69\% of the target locations with a similar base placement to SWS's home position. With an optimized base location, Baxter can reach 86\% of target locations. However, the optimized base location places the body of the robot in the middle of the clusters of points, which is infeasible for actual exercise with a robot.

\begin{figure*}[t]
    \centering
    \includegraphics[width=\textwidth]{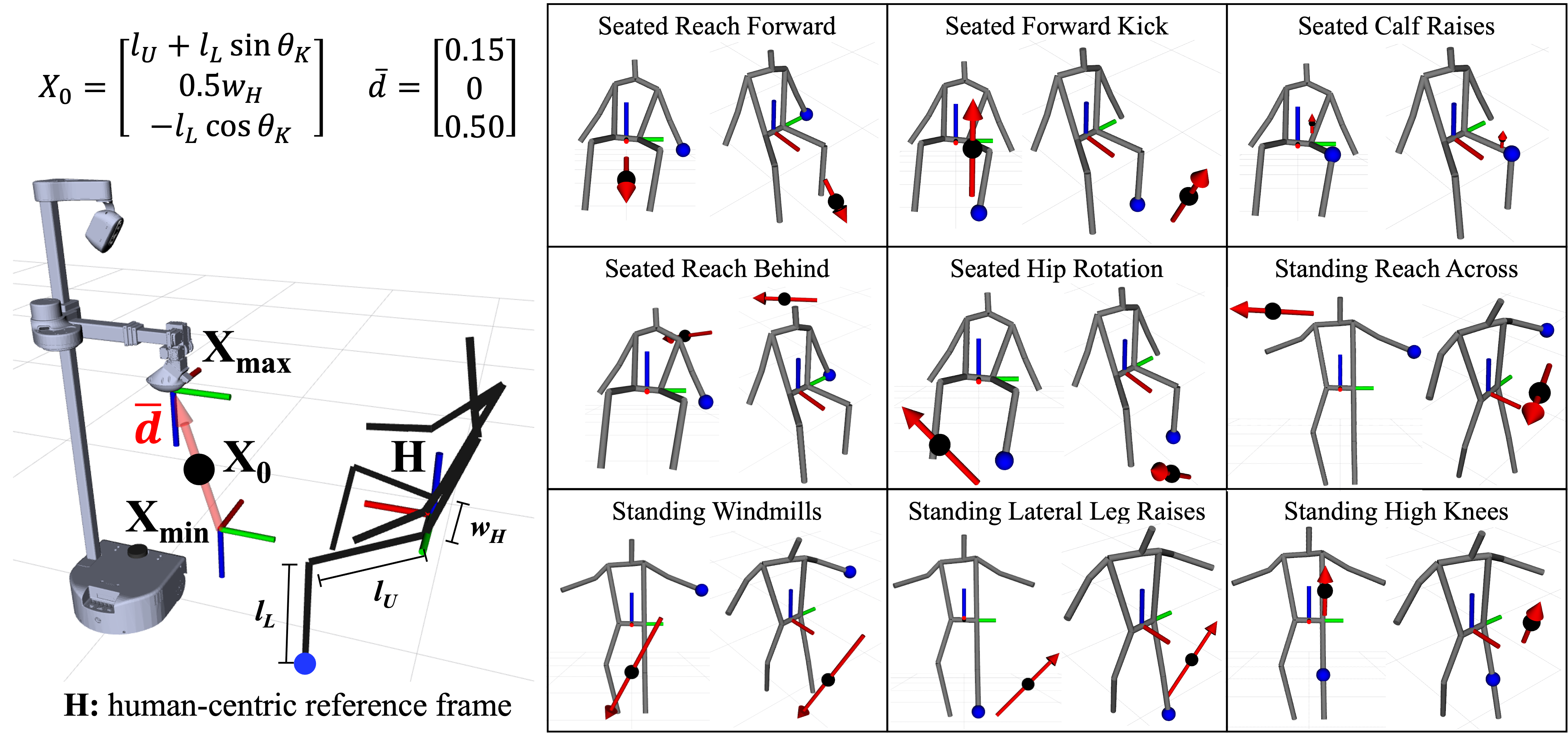}
    \caption{(\textbf{Left}) Exercise model for a seated forward kick. The location of the 50\% difficulty point $X_0$ is a function of relevant body dimensions (meters), and is shown as a black sphere. The difficulty vector $\Bar{d}$ is specified absolutely (meters), and is shown as a red arrow. The body part used to make contact with the robot during the exercise is indicated by a blue sphere. (\textbf{Right}) Front and isometric views of each of nine exercise models.} \vspace{-3mm}
    \label{fig:exercise_models}
\end{figure*}

SWS provides verbal instructions for each task using Google's \textit{gTTS} speech synthesizer \cite{pndurette}. These instructions are paired with a looping video demonstration of someone performing each task, shown on an external screen. SWS also provides feedback to users based on their exercise rate, and SWS informs users when it autonomously adjusts the difficulty of each exercise. 

To enrich human-robot interaction, SWS plays sound effects in reaction to events, such as chiming every time a user scores a point by touching the soft-bubble during an exercise. Other sound effects include playing tones with increasing in pitch while SWS "scans" a user's body dimensions with a pose estimator \cite{osokin2018real}, playing arpeggiated major chords to indicate task completion, and playing a buzzer sound with descending pitch to indicate when the user hits the bubble too hard.

\subsection{Personalized Exercise Target Locations}

SWS exercises involve the user repeatedly touching the robot's end effector (the exercise "target") with different parts of their body. The target location for each exercise varies between individuals, based on factors such as the length of their limbs or their range of motion. Considering these factors, SWS personalizes the location of the exercise targets using three sources of information:

\begin{enumerate}
    \item Personalized kinematic models of exercises based on an individual’s body dimensions provide an initial estimate of the target's appropriate location.
    \item Haptic sampling of an individual’s range of motion allows for direct estimation of the target's appropriate location.
    \item The target's location is dynamically adjusted based on models of a user's performance of an exercise.
\end{enumerate}

SWS uses simple kinematic models of each exercise to plan locations for the user to reach with their body parts. These models encourage participants to move their body parts through motions with a large amplitude. The models were formalized for deployment on a robot under the guidance of movement experts, who are investigators on the project and authors on the paper. \patrick{I don't think you can say movement experts, since that's not a credential, and you definitely can't say "co-investigators on this paper".} \matt{yeah, I'm not sure what would be appropriate here} \matt{update: uses Hackney's suggestion}



The models for nine of the ten exercises are shown in Figure \ref{fig:exercise_models}. The tenth exercise is Timed Up and Go \cite{shumway2000predicting}, where the robot moves 3m away from the user and measures the time taken for the user to touch its end effector. We define each exercise using three components. First, we define a 3D point that specifies a target (\textit{anchor point}) $X_0$ that requires a mild amount of stretching for a healthy adult to reach, as a function of an individual's body dimensions in meters. Second, we define a 3D difficulty vector $\Bar{d}$ that represents the direction for the target to move to increase the difficulty of the stretch, also in meters. Each of these components is specified relative to a coordinate system located between the user's hips, with the X axis pointing forwards and the Z axis pointing vertically against gravity. Lastly, we introduce a difficulty factor $F_{diff}$, where $0 \leq F_{diff} \leq 1$.

For a seated forward kick, 
\begin{equation}
    X_{0, kick} = \begin{bmatrix} l_U + l_L \sin \theta_k \\ 0.5 w_H \\ l_L \cos \theta_K \end{bmatrix} \;\;\;\;\;\;\;\; \Bar{d}_{kick} = \begin{bmatrix} 0.15 \\ 0 \\ 0.50 \end{bmatrix}
    \label{eqn:x0}
\end{equation}
where $l_U$ and $l_L$ are the length of the upper and lower leg, respectively, $w_H$ is the width of the hips, and $\theta_K$ is the knee extension angle from $90^\circ$. $X_0$ and $\Bar{d}$ are defined in a similar way for the other exercises. 

Together, $X_0$, $\Bar{d}$, and $F_{diff}$ allow for the definition of three points: $X_{min}$, the exercise target location with minimum difficulty; $X_{max}$, the exercise target location with maximum difficulty, and $X_{target}$, the personalized target location based on calibration of the $F_{diff}$ parameter. The relationships between these elements are given in (\ref{eqn:ex_target}).

\begin{equation}
    \begin{aligned}
        X_{min} &= X_0 - 0.5 \times \Bar{d} \\
        X_{max} &= X_0 + 0.5 \times \Bar{d} \\
        X_{target} &= X_{min} + F_{diff} \times \Bar{d}
    \end{aligned}
    \label{eqn:ex_target}
\end{equation}

Last, the robot adjusts the difficulty of the exercise between sets by modifying $F_{diff}$. The robot counts the number of repetitions performed, and determines whether the user's score falls into one of three brackets: \textit{Excellent}, \textit{Good}, or \textit{Poor}. These score brackets correspond to a set number of repetitions per minute (RPM) for each exercise. An \textit{Excellent} score results in increased difficulty, a \textit{Good} score results in no change to the difficulty, and a \textit{Poor} score results in decreased difficulty. These three ratings correspond to difficulty adjustments as specified in (\ref{eqn:delta}). We chose a difficulty adjustment of $\delta = 0.2$, corresponding to a 20\% difficulty shift along $\Bar{d}$. These adjustments are applied to $F_{diff}$ for the next set, as specified in (\ref{eqn:diff_adjust}).

\begin{equation}
    \begin{split}
    \Delta F_{diff} = \begin{cases}
        \delta  & \text{Excellent}\\
        0 & \text{Good} \\
        -\delta & \text{Poor}
    \end{cases} \\
    \end{split}
    \label{eqn:delta}
\end{equation}

\begin{equation}
\begin{split}
    F_{\mathit{diff (2)}} = F_{\mathit{diff (1)}} + \Delta F_{diff} \\
    X_{target_2} = X_{min} + F_{\mathit{diff (2)}} \times \Bar{d} \\
    \end{split}
    \label{eqn:diff_adjust}
\end{equation}

\textit{Algorithm \ref{alg:sws}} details the SWS exercise process. Here, the \textit{Exercise} method is invoked once for each of the exercises in the session. \textit{RunCalibrationMode} corresponds to the calibration procedure, \textit{RunPerformanceMode} corresponds to the repetitive exercise sets, and \textit{Performance Model} corresponds to the RPM score brackets for that exercise. The intent of this difficulty adjustment scheme is to challenge users without discouraging them. Combining range of motion measurements with adaptive difficulty models provides personalized exercise experiences, and opens the door for long-term tracking of exercise performance.

\begin{algorithm}
\caption{Adaptive Exercise Difficulty Algorithm: $E$ as ExerciseModel; $F_\mathit{diff}$ as difficulty factor; $B$ as user's Body Geometry; $X$ as target contact point}\label{alg:adpt-diff}
\begin{algorithmic}\\

\Function{Exercise}{$ $}
    \State $F_\mathit{diff} \gets RunCalibrationMode(B, E) $
    \While {not Done}
        \State $score \gets RunPerformanceMode(B, E, F_\mathit{diff}) $
        \State $\Delta F_\mathit{diff} \gets PerformanceModel(score)$
        \State $F_\mathit{diff} = F_\mathit{diff} + \Delta F_\mathit{diff}$
        \State Done $\gets CheckNumberOfSetsComplete()$
    \EndWhile
\EndFunction

\end{algorithmic}
\label{alg:sws}
\end{algorithm}

\subsection{Cognitive Exercise}

SWS also combines cognitive challenges with physical exercise. A common symptom of PD is difficulty with ``dual tasking” \cite{kalyani2019effects, smith2018communication}, or doing two things at the same time (e.g. walking while talking). By combining cognitive challenges with physical exercise, SWS can reinforce a PWP's dual tasking capabilities. SWS uses speech recognition to conduct category-specific naming exercises. For instance, SWS can prompt the participant to name distinct U.S. States or animals during an exercise. Participants say unique words of the specified category clearly and loudly into a wireless lapel mic each time they perform an exercise in repetition.


\section{User Study}

\subsection{Overview}

\begin{figure*}[t]
    \centering
    \includegraphics[width=0.9\textwidth]{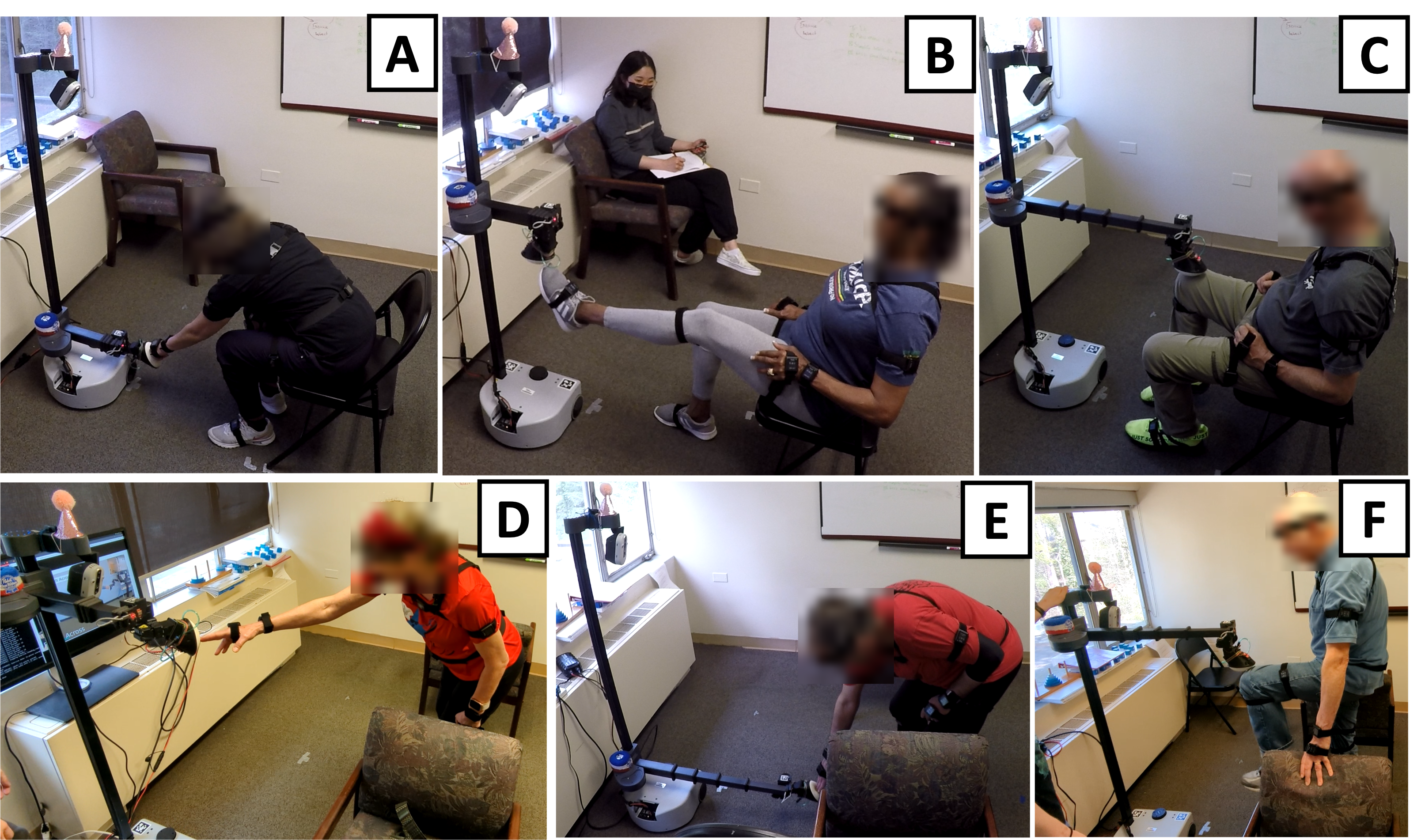}
    \caption{We performed a user study with 10 PWP. (\textbf{A-C}) Participants performing seated reaching, kicking, and calf raise exercises. (\textbf{D-F)} Participants performing standing reach across, reach down, and high knees exercises. Chairs were placed near the participant as safety devices for the standing exercises. \youliang{Using half a space for this figure is abit wasteful}}
    \label{fig:participants}
\end{figure*}
We performed a preliminary user study with 10 PWP. This study was approved by Institutional Review Boards at Georgia Institute of Technology (protocol H22359) and the Emory University School of Medicine (protocol STUDY00004909) \patrick{typically you just say "was approved by an Institutional Review Board}. All participants provided written informed consent prior to participation. \matt{updated based on Hackney's feedback} Participants were recruited for one session, lasting for a total of about 2.5 hours. First, participants performed several motor and cognitive pre-screening assessments under the guidance of a researcher. Then, participants were introduced to the robot. Next, the robot led each participant through an exercise session consisting of three seated exercises and three standing exercises. Two of the exercises also contained a cognitive component, as shown in Table \ref{tab:exercise_sequence}. The full exercise session lasted for approximately one hour, and participants were allowed to take breaks at any time throughout the session.

\begin{table}
    \centering
    \begin{tabular}{|c|c|c|c|} \hline
        Exercise & Exercise & Cognitive & In-Contact \\
        Number & Name &  & Body Part \\ \hline
        1 & Seated Reach Forward & None & Hand \\ 
        2 & Seated Forward Kick & None & Foot \\ 
        3 & Seated Calf Raises & U.S. States & Knee \\ 
        4 & Standing Reach Across & Animals & Hand \\ 
        5 & Seated Windmills & None & Hand \\ 
        6 & Seated High Knees & None & Knee \\ \hline
    \end{tabular}
    \caption{User Study Exercise Sequence}
    \label{tab:exercise_sequence}
\end{table}

Each exercise was performed as follows. First, the robot provided instructions for running the calibration sequence. Then, the user calibrated their right and left ranges of motion. Next, the robot provided instructions for the actual exercise. Last, participants performed two 30-second sets of the exercise with their right-side body part, followed by two 30-second sets with their left-side body part. The robot adjusted the difficulty of the exercise between each set of the exercise according to (\ref{eqn:delta}). After the exercise session, participants were administered a series of questionnaires regarding their experience.

\subsection{Subjective Measures}

One of the administered questionnaires was an adapted version of the Robot Opinions Questionnaire developed by Chen, et. al. \cite{chen2017older}. This survey contains 22 Likert type questions (5-point scale, 1 = strongly disagree, 5 = strongly agree) that assess five attitudes related to Technology Acceptance Model (TAM) \cite{venkatesh2003user} literature: Perceived Usefulness (PU), Perceived Ease of Use (PEOU), Perceived Enjoyment (PENJ), Intent to Use (ITU), and Positive Attitudes (ATT).



Additionally, we asked each participant to provide Ratings of Perceived Exertion (RPE) at three points during the trial: first, before exercise began; second, between the seated calf raises exercise and standing reach across exercise; and last, after the standing high knees exercise. Participants were instructed to rate their perceived physical effort on a scale from 1-10, with 1 corresponding to sitting on the couch, and 10 corresponding to having just run a marathon, which is similar to the Borg CR-10 RPE scale \cite{borg1998borg}.





\section{Results and Discussion}

Participants from the study are shown performing each of the exercises in Figure \ref{fig:participants}. We recruited 10 participants (5 male, 5 female) aged 69$\pm$5.7 years who self-reported having PD with a severity between 1$-$2.5 on the modified Hoehn and Yahr scale \cite{goetz2004movement}.

\subsection{Subjective Measures}

\begin{figure}
    \centering
    \includegraphics[width=0.5\textwidth]{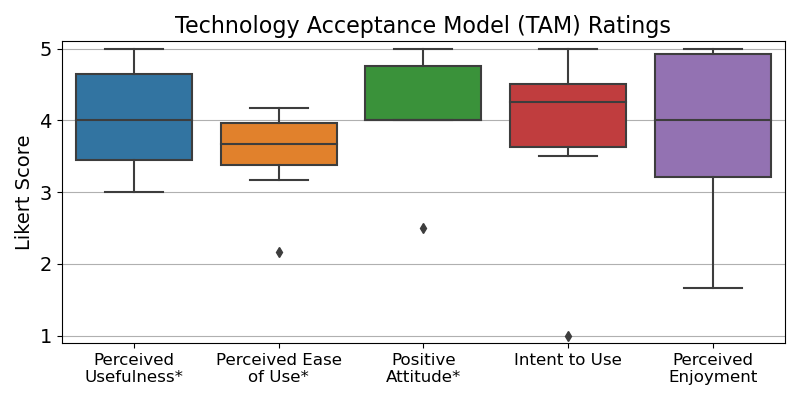}
    \caption{Robot Opinions Questionnaire survey response distributions.} \vspace{-3mm}
    \label{fig:tam}
\end{figure}

Cronbach's $\alpha$ values for each TAM attitude ranged between 0.88-0.98, which indicates that the responses were internally consistent. This allows us to compute an average response across questions for each attitude for each participant \cite{chen2017older}. Box plots of these averages are shown in Figure \ref{fig:tam}, and summary statistics are given in Table \ref{tab:tam}.

For each attitude, we performed a two-tailed Wilcoxon signed-rank test to determine whether the response distribution differs from a neutral response (3 out of 5 on a 5-point scale). We found that PU ($p=0.0039$), PEOU ($p=0.037$), and ATT ($p=0.039)$ were all rated as significantly positive ($p_{crit} = 0.05$). However, ITU ($p=0.064$) and PENJ ($p=0.055$) were not significantly positive. 


\begin{table}
    \centering
    \begin{tabular}{|c|c|c|c|c|c|} \hline
        Attitude & $\alpha$ & Median & Range & \textit{p}-value & W-stat \\ \hline
        PU* & 0.94 & 4.0 & 2.0 & \textbf{0.004} & 0.0 \\ \hline
        PEOU* & 0.88 & 3.7 & 2.0 & \textbf{0.037} & 6.5 \\ \hline
        ATT* & 0.96 & 4.0 & 2.5 & \textbf{0.039} & 1.0 \\ \hline
        ITU & 0.92 & 4.25 & 4.0 & 0.064 & 9.0 \\ \hline
        PENJ & 0.98 & 4.0 & 3.3 & 0.055 & 6.0 \\ \hline
    \end{tabular}
    \caption{Summary statistics for the Robot Opinions Questionnaire survey responses. *indicates statistical significance, \textbf{$p_{crit} = 0.05$} \patrick{just because you have all the stats doesn't mean you need to show them. I would leave out all the stats except the important ones. IMO, range, W-stat, and alpha can go.}}
    \label{tab:tam}
\end{table}

\begin{figure}
    \centering
    \includegraphics[width=0.5\textwidth]{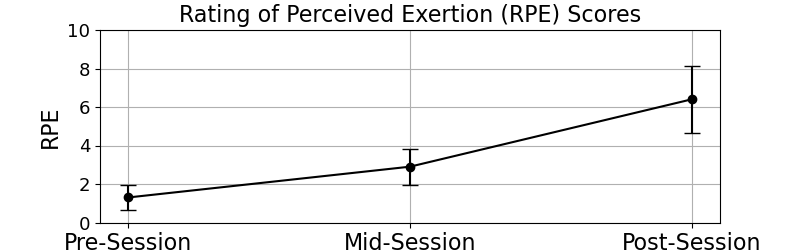}
    \caption{RPE reported by participants before, halfway through, and after the SWS exercise protocol. \patrick{This is not the right plot for the job. A line plot with the x-axis being pre/mid/post, including error bars, would be much more effective.}} \vspace{-3mm}
    \label{fig:rpe}
\end{figure}

As shown in Figure \ref{fig:rpe}, participants reported an average RPE of 1.3$\pm$0.64 before, 2.9$\pm$0.94 halfway through, and 6.4$\pm$1.7 immediately after the exercise session, which shows a clear increase in perceived effort across the session. \cite{salgado2013evidence}.





\subsection{Discussion}

Our results provide evidence that Stretch with Stretch (SWS) can provide personalized physical therapy for people with Parkinson's disease (PWP). SWS led participants through six different exercises in which participants used their hands, knees, and feet to repeatedly make contact with the robot. Participants also played cognitive games in conjunction with two of the exercises. Participants tended to perceive SWS as useful and easy to use and to report that they had positive attitudes. Participants' average ratings of perceived exertion were consistent with recommendations for people with mild to moderate Parkinson's disease \cite{salgado2013evidence}. 

SWS's size and weight make it feasible for use by individuals in homes. Fully realizing this potential will require further development. For example, due to its mobility, SWS has the potential to proactively seek out users and engage them in exercise, which might improve adherence. SWS does not currently observe or provide feedback on exercise form, which varied across participants. Also, enabling SWS to guide a person with Parkinson's disease through exercises without experimenters present would require careful consideration of safety. 

Overall, our work suggests that SWS is a promising direction for rehabilitation robots and use in clinics with supervision may be achievable in the near term.

\section{Conclusion}

In this paper we presented Stretch with Stretch (SWS), a system for personalized robot-led physical therapy for people with Parkinson's disease (PWP). The system uses a novel model for personalized exercise that adapts to an individual's kinematics, range of motion, and performance. Results from a study we conducted with PWP suggests that SWS is a promising approach for physical therapy. 





\section*{ACKNOWLEDGEMENTS}

We thank the Toyota Research Institute and the McCamish Parkinson's Disease Innovation Program for funding this research. We thank Jose Barreiros, Calder Phillips-Grafflin, Patrick Grady, and Jeremy Collins for their feedback and discussions about this project.



\bibliographystyle{ieeetr} 
\bibliography{ref}  

\end{document}